\documentclass[conference,a4paper]{IEEEtran}
\IEEEoverridecommandlockouts
\usepackage{cite}
\usepackage{amsmath,amssymb,amsfonts}
\usepackage{algorithmic}
\usepackage{graphicx}
\usepackage{textcomp}
\usepackage{xcolor}
\usepackage{multirow}
\usepackage{booktabs}
\usepackage{subcaption}
\usepackage{array}
\usepackage{float}
\usepackage{bbding}
\usepackage{stfloats}

\def\BibTeX{{\rm B\kern-.05em{\sc i\kern-.025em b}\kern-.08em
    T\kern-.1667em\lower.7ex\hbox{E}\kern-.125emX}}

\makeatletter

\def\ps@IEEEtitlepagestyle{%
  \def\@oddfoot{\mycopyrightnotice}%
  \def\@evenfoot{}%
}
\def\mycopyrightnotice{%
  {\footnotesize 979-8-3503-7903-7/24/\$31.00 \textcopyright 2024 IEEE \hfill}% <--- Change here
  \gdef\mycopyrightnotice{}% just in case
}

\begin{document}

\title{DeepIcon: A Hierarchical Network for Layer-wise Icon Vectorization}
\author{\IEEEauthorblockN{Qi Bing,
Chaoyi Zhang, and Weidong Cai}
\IEEEauthorblockA{\textit{School of Computer Science} \\
\textit{The University of Sydney}\\
NSW 2006, Australia \\
qbin4920@uni.sydney.edu.au, chaoyivision@gmail.com, tom.cai@sydney.edu.au}}

\maketitle
% \IEEEpubidadjcol
% \thispagestylefirstpage

\begin{abstract}
In contrast to the well-established technique of rasterization, vectorization of images poses a significant challenge in the field of computer graphics. Recent learning-based methods for converting raster images to vector formats frequently suffer from incomplete shapes, redundant path prediction, and a lack of accuracy in preserving the semantics of the original content. These shortcomings severely hinder the utility of these methods for further editing and manipulation of images. To address these challenges, we present DeepIcon, a novel hierarchical image vectorization network specifically tailored for generating variable-length icon vector graphics based on the raster image input. Our experimental results indicate that DeepIcon can efficiently produce Scalable Vector Graphics (SVGs) directly from raster images, bypassing the need for a differentiable rasterizer while also demonstrating a profound understanding of the image contents. 
\end{abstract}


\begin{figure*}[!hb]
  \centering
  \begin{subfigure}{0.28\linewidth}
    \centering
    % \fbox{\rule{0pt}{0.5in} \rule{.9\linewidth}{0pt}}
    \includegraphics[width=1\linewidth]{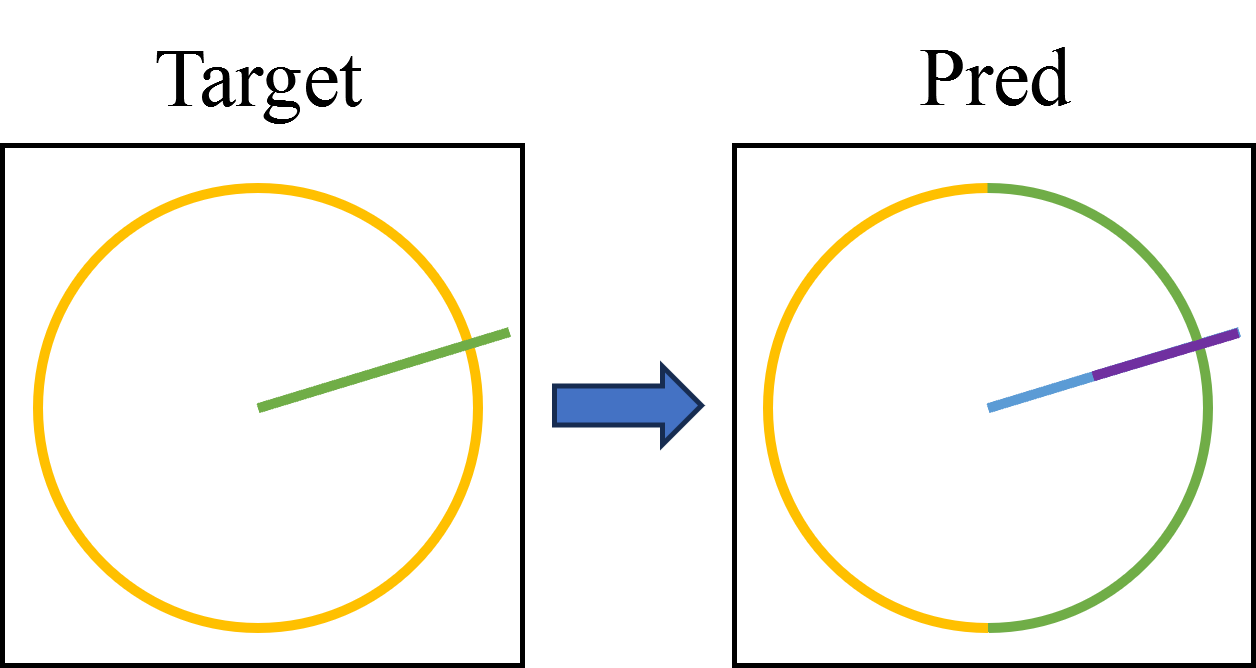}
    \caption{}
    \label{fig:example-a}
  \end{subfigure}
  \hfill
  \begin{subfigure}{0.28\linewidth}
    \centering
    \includegraphics[width=1\linewidth]{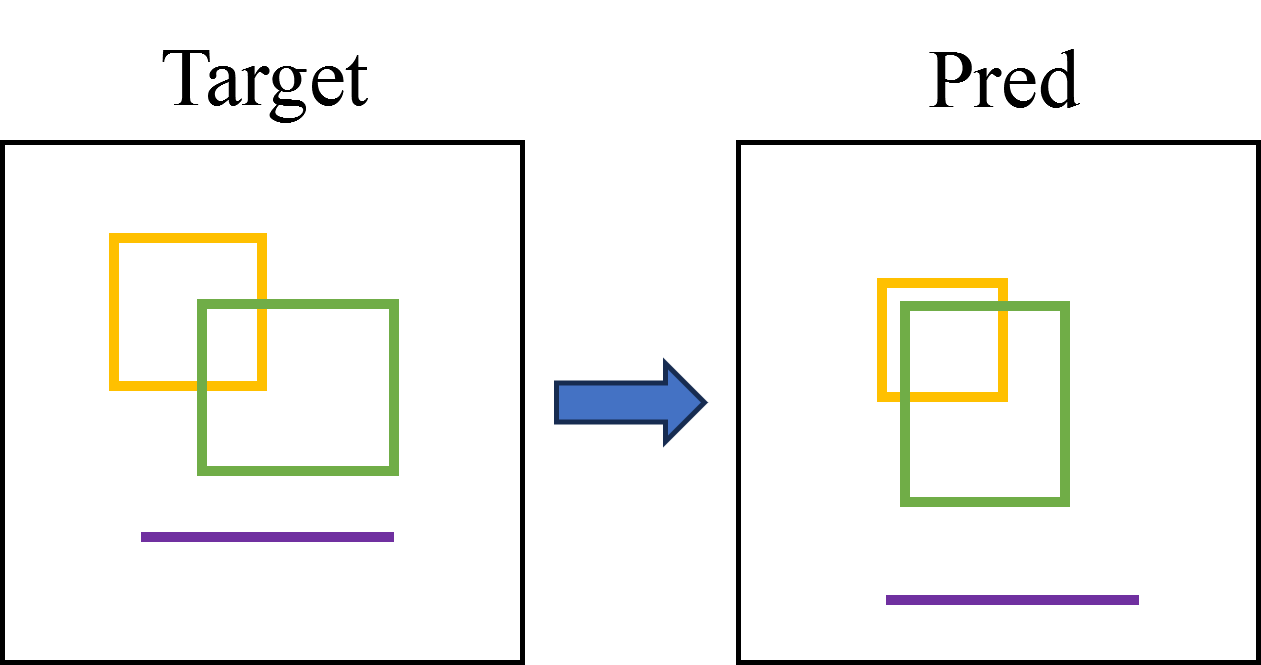}
    % \fbox{\rule{0pt}{0.5in} \rule{.9\linewidth}{0pt}}
    \caption{}
    \label{fig:example-b}
  \end{subfigure}
  \hfill
  \begin{subfigure}{0.28\linewidth}
    \centering
    \includegraphics[width=1\linewidth]{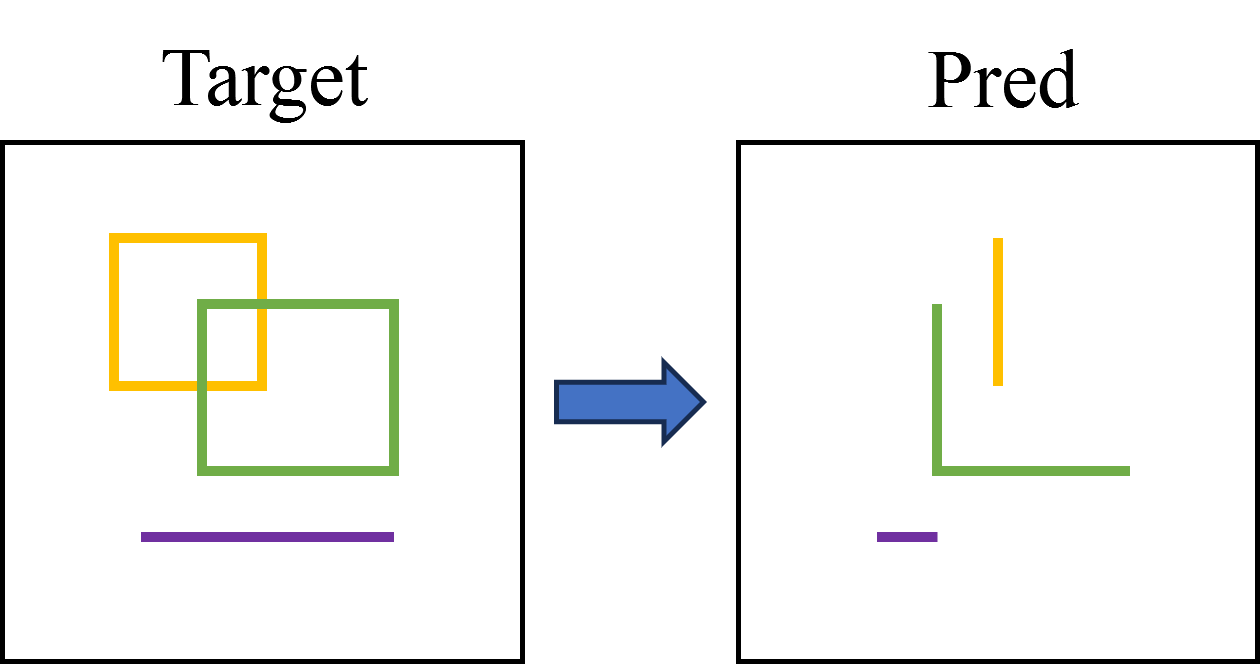}
    % \fbox{\rule{0pt}{0.5in} \rule{.9\linewidth}{0pt}}
    \caption{}
    \label{fig:example-c}
  \end{subfigure}
  \caption{Three examples of frequently encountered issues in image vectorization. The colors in the images are used to distinguish individual SVG paths. (a): An example of redundant path predictions. Despite the redundancy, the rendered output remains quantitatively accurate, indicating that redundant paths may not be reflected through the evaluation metrics. (b): In this case, the prediction performance results in low quantitative accuracy due to the geometric offset of the predicted paths. However, this example demonstrates the model's capability to grasp the underlying semantics and relationships between shapes, such as recognizing two rectangles and one line. (c): Compared with (b), predicting shapes incompletely but with accurate positioning may achieve higher pixel accuracy.}
  \label{fig:example}
\end{figure*}
% \begin{figure*}[!hb]
%   \centering
%   \includegraphics[width=1\linewidth]{pics/overview.png}
%   \caption{The fundamental workflow that DeepIcon performs image vectorization, converting images into SVG format. Initially, the image undergoes encoding to transform into a single embedding. This embedding is then fed into a decoder to generate a sequence of parametric shapes.
%   }
%   \label{fig:overview}
% \end{figure*}

\section{Introduction}
\label{sec:intro}
\begin{figure*}[tb]
  \centering
  \includegraphics[width=1\linewidth]{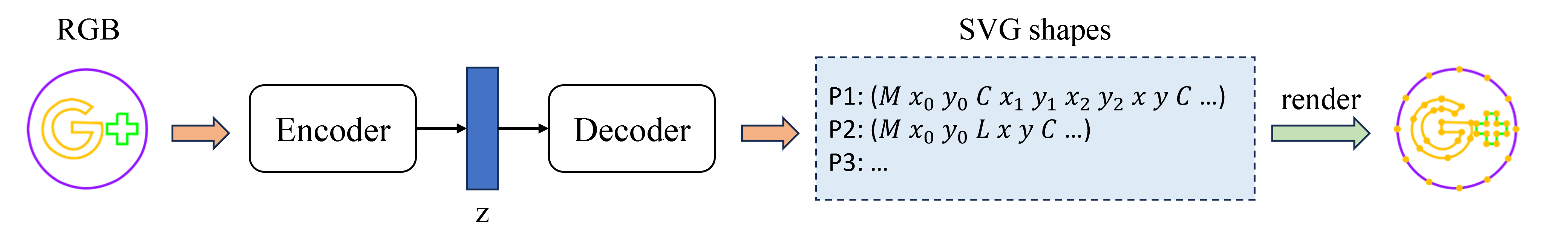}
  \caption{The fundamental workflow that DeepIcon performs image vectorization, converting images into SVG format. Initially, the image undergoes encoding to transform into a single embedding. This embedding is then fed into a decoder to generate a sequence of parametric shapes.
  }
  \label{fig:overview}
\end{figure*}
% \begin{figure*}[tb]
%   \centering
%   \begin{subfigure}{0.28\linewidth}
%     \centering
%     % \fbox{\rule{0pt}{0.5in} \rule{.9\linewidth}{0pt}}
%     \includegraphics[width=1\linewidth]{pics/example_a.png}
%     \caption{}
%     \label{fig:example-a}
%   \end{subfigure}
%   \hfill
%   \begin{subfigure}{0.28\linewidth}
%     \centering
%     \includegraphics[width=1\linewidth]{pics/example_b.png}
%     % \fbox{\rule{0pt}{0.5in} \rule{.9\linewidth}{0pt}}
%     \caption{}
%     \label{fig:example-b}
%   \end{subfigure}
%   \hfill
%   \begin{subfigure}{0.28\linewidth}
%     \centering
%     \includegraphics[width=1\linewidth]{pics/example_c.png}
%     % \fbox{\rule{0pt}{0.5in} \rule{.9\linewidth}{0pt}}
%     \caption{}
%     \label{fig:example-c}
%   \end{subfigure}
%   \caption{Three examples of frequently encountered issues in image vectorization. The colors in the images are used to distinguish individual SVG paths. (a): An example of redundant path predictions. Despite the redundancy, the rendered output remains quantitatively accurate, indicating that redundant paths may not be reflected through the evaluation metrics. (b): In this case, the prediction performance results in low quantitative accuracy due to the geometric offset of the predicted paths. However, this example demonstrates the model's capability to grasp the underlying semantics and relationships between shapes, such as recognizing two rectangles and one line. (c): Compared with (b), predicting shapes incompletely but with accurate positioning may achieve higher pixel accuracy.}
%   \label{fig:example}
% \end{figure*}

Recent research in the field of computer graphics has predominantly focused on data processing and the interpretation of raster images. In contrast, vector graphics have attracted relatively less attention within the deep learning community. Despite this, vector graphics have emerged as a preferred format across various domains, including web graphics, user interfaces, animation, and icon design, owing to their adaptability in web applications and superior scalability. Unlike raster graphics, vector graphics offer enhanced precision in shape representation at varying resolutions and encapsulated higher-level information, such as the number of segments, sizes of segments, and the spatial relationships among paths. This inherent complexity provides significant advantages over raster graphics in many applications but poses considerable challenges for learning-based methods regarding representation, comprehension, and generation of vector graphics. In this paper, we concentrate on Scalable Vector Graphics (SVG), one of the most popular and widely utilized vector graphics formats, to explore the generation of vector graphics from images.

Image rasterization is a relatively straightforward process, whereas reversing an image to its vector representation presents significant challenges, primarily due to the potential non-uniqueness of results. Also, traditional methods perform poorly in accurately interpreting and preserving the semantics and topological characteristics of image content. In contrast, learning-based approaches have emerged as more promising in recent research, offering solutions to prevailing issues. Nevertheless, despite the advancements made by these novel approaches to image vectorization, they frequently encounter three significant obstacles: incomplete shape prediction as shown in Fig.~\ref{fig:example-c}, redundant shape predictions as shown in Fig.~\ref{fig:example-a}, and the inherent limitation of using pixel loss as a metric to evaluate the performance of vector graphics generation as shown in Fig.~\ref{fig:example-b}. Low processing speed can make vectorization impractical for real-time applications or large-scale processing tasks. On the other hand, unnecessary complexity in layer segmentation can result in vectorized images that are difficult to edit or manipulate, limiting their utility for graphic design and professional artwork. Recent studies~\cite{LIVE, DiffVG, PaintTF, StylizedNeural} have conducted image vectorization by optimizing a fixed number of initialized parametric strokes, heavily relying on the use of differentiable rasterizers for loss backpropagation. As previously demonstrated, the reliance on pixel loss for backpropagation is suboptimal for vector graphics generation, often resulting in convergence to local optima or redundant predictions. Although these methods achieve visually appealing outcomes, they often neglect semantic information, leading to a tendency towards redundant shapes. Conversely, approaches such as~\cite{SVG-VAE, DeepSVG}, which utilize Variational Autoencoders (VAEs) for SVG reconstruction, excel at preserving topological characteristics in vector generation. However, they failed to reach accurate shapes.

Different from representing image as a grid of pixels, SVG conceptualize images as sequences of parametric 2D shapes. These shapes comprise lists of lines or curves defined by specific points (or coordinates). Owing to its sequential and hierarchical nature, SVG can be simplified and represented as an extended sequence, enabling the application of recent advancements in Natural Language Processing (NLP), such as Recurrent Neural Networks (RNNs), Transformer-based models \cite{transformer}, Generative Pre-trained Transformer (GPT-2 \cite{gpt3}), and Multimodal NLP. Drawing inspiration from the hierarchical SVG generation pipeline utilized in DeepSVG \cite{DeepSVG}, we introduce DeepIcon, a novel image vectorization network. DeepIcon is designed to hierarchically generate the final SVG from a single image input without depending on a differentiable rasterizer. Different from DeepSVG, we use a CLIP-based image encoder instead of a hierarchical SVG encoder. Also, we process the output SVG paths in the continuous space. Our experimental results demonstrate that DeepIcon outperforms state-of-the-art optimization-based vectorization approaches in preserving the topology similarity and geometric accuracy.

In summary, we have designed a novel hierarchical SVG generation model that converts an input RGB image into a high-quality SVG representation. Our contributions are threefold:
\begin{itemize}
  \item 
  % \BQ{Define a complex graphics program.}
  As illustrated in Fig.~\ref{fig:overview}, we developed an image vectorization pipeline named DeepIcon that facilitates high-quality image generation from images to Scalable Vector Graphics (SVG). This innovative pipeline not only achieves superior IMG-to-SVG translation fidelity but also ensures the preservation of the intrinsic geometric and relational information.
  \item 
  % \BQ{Novel pipeline for program generation.}
  We propose an accurate SVG decoder capable of generating a multi-path SVG result from a single embedding, denoted as $z$. Furthermore, we validate the versatility of the proposed decoder by demonstrating its applicability to various types of data inputs, provided that a corresponding encoder is available. 
  \item 
  % \BQ{Result surpasses SOTA methods.}
  We evaluate our method with a comparison to current state-of-the-art image vectorization and SVG generation approaches. The experimental results conclusively demonstrate that DeepIcon outperforms these methods both quantitatively and qualitatively.
\end{itemize}

\section{Related Works}
\subsection{Image Vectorization}
Despite the straightforwardness of image rasterization from vectors, image vectorization from raster images is much more difficult due to the inadequate information provided in the image. Critical details such as the order of overlapping shapes, the direction in which a shape should be drawn, and the segmentation of shapes into distinct entities are not explicitly defined in raster images. While humans can often interpret these aspects intuitively, they pose considerable challenges for computational approaches, especially when dealing with complex images. In response to these challenges, learning-based approaches have been increasingly promising due to their efficiency in extracting higher-level information from images. 

To bridge the gap between parametric attributes and image data, recent studies \cite{LIVE, DiffVG, PaintTF, StylizedNeural, clipasso, geometry_sketch} adopt differentiable rasterizers to directly backpropagate pixel losses and optimize the parameters with consideration for the rasterized output, facilitating vector graphic generation without the need for ground-truth vector supervision. Moreover, Cloud2Curve \cite{Das2021Cloud2CurveGA} approaches the generation of parametric curves autoregressively from hand-drawn sketches by initially processing these sketches into a sampled point cloud. Despite their innovative approach, reliance on differentiable rasterizers has shown its weakness of redundant shapes and convergence to local optima, underscoring a critical limitation in preserving topological integrity. An alternative approach conceptualizes vector graphics as executable graphics programs. Studies \cite{NEURIPS2019_50d2d226,Sharma_2018_CVPR, Feser2023InductivePS, lego, EllisRST18, my_icpr} have introduced their self-defined domain-specific language (DSL) that interprets vector graphics as compositions of various parametric primitives (e.g., circles, rectangles, straight lines and Lego bricks). There are also attempts such as \cite{ClipGen, SAMVGAM, dualvector} that incorporate additional knowledge (e.g., category, segmentation masks, and contours) to aid the vectorization process. 

Unlike most existing vectorization approaches, our method does not rely on using a differentiable rasterizer or providing additional information. Operating solely with a single image input, DeepIcon can generate high-quality SVG representations featuring a variable number of shapes.
\begin{table*}[t]
\begin{center}
\caption{Description of SVG commands. By default, we set the start point for each path (a sequence of commands) as ($0$, $0$). The default start point for each command is the end point of the last command in the path.}
\vspace*{1mm}
\label{tab:SVGs}
\begin{tabular}{ c  c  p{0.19\linewidth}  c }
  \hline
  Command & Arguments & \multicolumn{1}{c}{Description} & Example \\ \hline
  M & \multirow{2}{*}{$x$, $y$} & Move the current cursor to  & \
      \multirow{2}{*}{\includegraphics[scale=0.25]{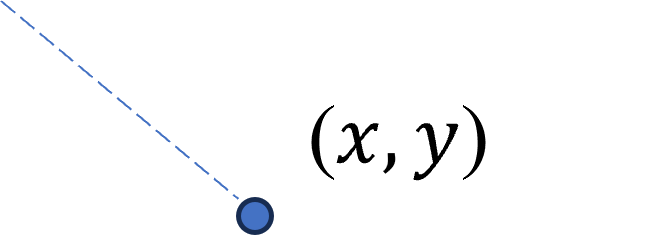}}\\ 
  (MoveTo) & & point ($x$, $y$).\\ \hline

  L & \multirow{3}{*}{$x$, $y$} & Draw a straight line from the   & \
      \multirow{3}{*}{\includegraphics[scale=0.25]{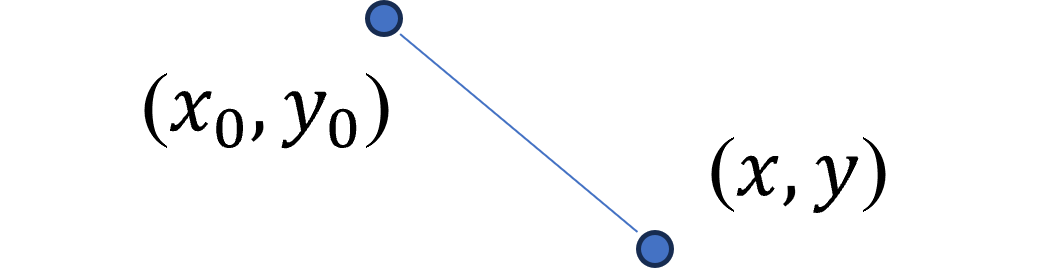}}\\ 
  (LineTo) & & current point ($x_0$, $y_0$) to point \\ 
  & & ($x$, $y$). \\ \hline

  \multirow{1.5}{*}{C} & $x_1$, $y_1$ & Draw a cubic bézier curve  &
      \multirow{3}{*}{\includegraphics[scale=0.25]{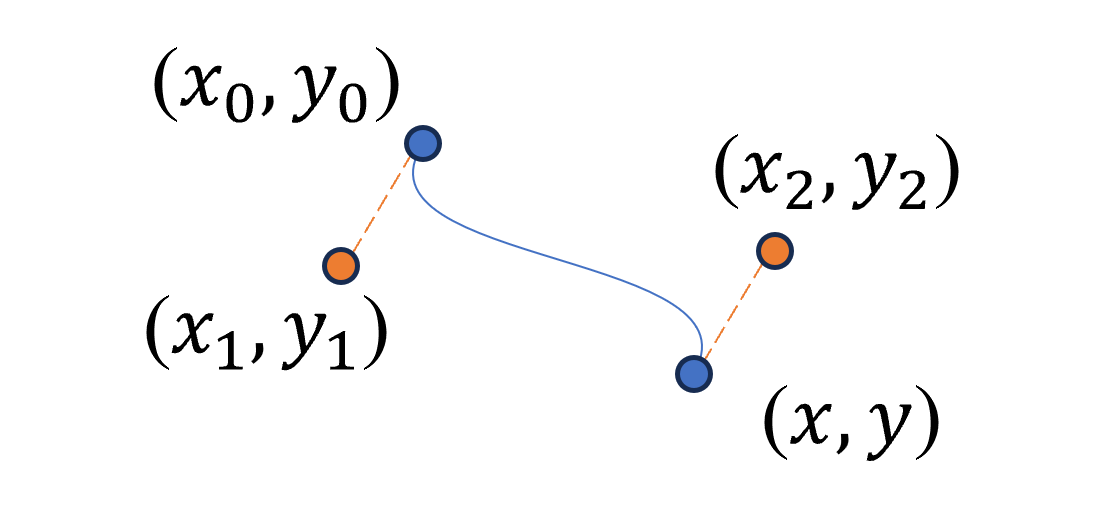}}\\ 
  \multirow{1.5}{*}{(Cubic Bézier)} & $x_2$, $y_2$ & to point ($x$, $y$) with control\\ 
  & $x$, $y$ & points ($x_1$, $y_1$) and ($x_2$, $y_2$).\\ \hline
\end{tabular}
\end{center}
\end{table*}

\subsection{Vector Graphics Generation}
Existing methods have proposed different vector representations with different aspects of simplifications. With the presentation of sketch datasets `Quick, Draw!' \cite{quickdraw}, Sketch-RNN emerged as the first Sequence-to-Sequence Variational Autoencoder (VAE) model conducting conditional sketch generation and interpolation. Inspired by the success of Sketch-RNN, subsequent researchers proposed improved sketch generation approaches by leveraging the recent improvements in the Natural Language Processing (NLP) domain. For instance, Sketchformer~\cite{Sketchformer2020} introduced a shift from LSTM to a Transformer-based architecture, achieving significant improvements in interpolation stability over Sketch-RNN. Similarly, \cite{tang2021write, Dai2023DisentanglingWA} focuses on the generation of stylized handwritten Chinese characters by disentangling style and content using two RNN-based or transformer-based encoders. 

Sketches conceptualize vector graphics as a sequence of densely plotted points accompanied by corresponding pen states, which makes it flexible to represent a wide range of images. However, there is still a gap between such sketch representations and the widely adopted vector graphics formats (e.g., SVG). Some of the recent studies focus on generating complex vector graphics that incorporate both straight lines and curves, thereby introducing an inherent hierarchical structure to the sequential representation. Notably, \cite{SVG-VAE, Aoki2022SVGVF, SVGformer} proposed generative models specifically for SVG vector fonts, while \cite{DeepSVG} has concentrated on learning to generate SVG icons from a variational auto-encoder (VAE) based structure. Also, DeepSVG \cite{DeepSVG} makes an icon dataset public, providing the processed SVG annotations. \cite{Wu2023IconShopTV} proposed methods for generating text-conditioned SVG icons using an autoregressive transformer, achieving high quality icon generation results. 

Furthermore, there is also recent research focusing on the generation of graphics layouts. Similar to SVGs, graphics layouts are also composed of primitive types (e.g., topic categories and element types) and parametric arguments. \cite{LayoutVAE, Zheng2019ContentawareGM, Lee2019NeuralDN} attempt conditional layout generation with different types of conditions provided (e.g., label sets, metadata, partial elements, and relation annotations). \cite{CanvasVAE} learns to generate unconditional vector graphic documents through a transformer-based VAE structure.

Our proposed novel SVG generation network is inspired by DeepSVG \cite{DeepSVG}, addressing the problem of poor geometric accuracy and improving the visual results by using autoregressive path decoders and predicting arguments in the continuous space.

\section{DeepIcon}
In this section, we first introduce the definition of our SVG representation (Section~\ref{sec:svg}) and the strategy for tokenization (Section~\ref{sec:tokenization}), followed by a comprehensive description to our proposed network (Section~\ref{sec:archi}). We then proceed to detail the implementation specifics of the training process (Section~\ref{sec:training}).

\subsection{SVG Representation}
\label{sec:svg}
As a widely adopted representation, SVG offers a large range of commands and functionalities for designers to work with. To limit the scope for training and SVG inference, we simplified the representation of SVG by only picking the necessary attributes for icon representations. Here we present the SVG commands we choose to utilize in this paper as referenced in Table~\ref{tab:SVGs}. Since most basic shapes (e.g., rectangle, circle, ellipse, line, polyline and polygon) can be represented by a composition of straight lines and curves, with these three commands listed in the table, our model is capable of constructing a variety of shapes as through combinations of curves and lines. For instance, rather than employing the specialized $Rect$ command for rectangles, we opt for a more fundamental approach: initiating with the $M(x, y)$ command to move the cursor to a starting position, followed by the execution of four straight lines using the $L(x, y)$ command. By default, we set the initial point of each path at $(0, 0)$. Moreover, we simplify the command set by removing all attributes despite geometric arguments (coordinates of control points). Consequently, an SVG script $S$ is succinctly represented as $S=\{P_i\}$, where $P_i$ denotes the $i^{th}$ path within the SVG script. Each path $P_i=\{C_i^j\}$ comprises a sequence of commands, where $C_i^j$ denotes $j^{th}$ command within $P_i$. $N_P$ and $N_C$ denote the  number of paths and commands in each path respectively. Examples of the command implementations $C_i^j$ are provided in Table~\ref{tab:SVGs}. 

\subsection{Tokenization}
\label{sec:tokenization}
Different from most discrete tokenization methods, our approach to SVG decoding emphasizes enhanced geometric accuracy by treating point positions as continuous arguments. 
% \CY{Specifically, we tokenize each command $C_i^j = (\{T_k\}, \{A_k\})$ with a pair of sequences (with $i$ and $j$ omitted), where $T_k$ and $A_k$ denote the $k^{th}$ elements of the type sequence $T$ and args sequence $A$, respectively.}
% \CY{In detail, the ${T_k}$ sequences describe the discrete token types with $T_k\in \{\mathtt{M}, \mathtt{L}, \mathtt{C}, \mathtt{arg}, \mathtt{SOS}, \mathtt{EOS}\}$, while ${A_k}$ sequence indicates the continuous location arguments.}
To facilitate this, we represent a single command implementation as $C_i^j=(\{T_k\}, \{A_k\})$ with a pair of sequences (with $i$ and $j$ omitted), where $T_k$ and $A_k$ denote the $k^{th}$ elements of the type sequence $T$ and argument sequence $A$, respectively. In detail, the $\{T_k\}$ sequence describes the discrete token types with $T_k\in \{\mathtt{M}, \mathtt{L}, \mathtt{C}, \mathtt{arg}, \mathtt{SOS}, \mathtt{EOS}\}$, while $\{A_k\}$ sequence indicates the continuous location arguments.
These discrete tokens $T_k$ respectively specify the token types MoveTo, LineTo, Cubic Bézier, continuous arguments, the start of SVG, and the end of SVG. 
% The true value at position with an `ARG' within the type sequence $U_i^j$ is picked from the same position at $A_i^j$.
Furthermore, to standardize the length of sequences for each command $C_i^j$, we also implement padding with a value of $-1$ for sequences $T$ and $A$. 
% For instance, commands such as $M(x,y)$ and $L(x,y)$ undergo padding to become $M(-1, -1, -1, -1, x, y)$ and $L(-1, -1, -1, -1, x, y)$, respectively, ensuring uniform sequence lengths across all command types. The command $C(x_1, y_1, x_2, y_2, x, y)$ remains unaltered, where its pair of sequence would be $U_i^j=[$C, `ARG', ..., `ARG'$]$ and $A_i^j=$[$-1$, $x_1$, $y_1$, $x_2$, $y_2$, $x$, $y$] with unused values in $A_i^j$ set to $-1$. 

Sequences of commands within the same path $P_i$ are then concatenated into a single pair of extended sequences for subsequent model processing. Consequently, the final tokenized representation of a path $P_i$ is formulated as $P_i=(\{T_{j,k}\}, \{A_{j,k}\})$ (with $i$ omitted), where $\{T_{j,k}\}$ and $\{A_{j,k}\}$ indicate the concatenated sequence of tokens for types and arguments within the $i^{th}$ path, ensuring a coherent and standardized structure for autoregressive decoding. 
During training, the special tokens $\mathtt{SOS}$ and $\mathtt{EOS}$ are added to the start and end of type sequence $\{T_{j,k}\}$ respectively.   

\begin{figure*}[tb]
  \centering
  \includegraphics[width=0.7\linewidth]{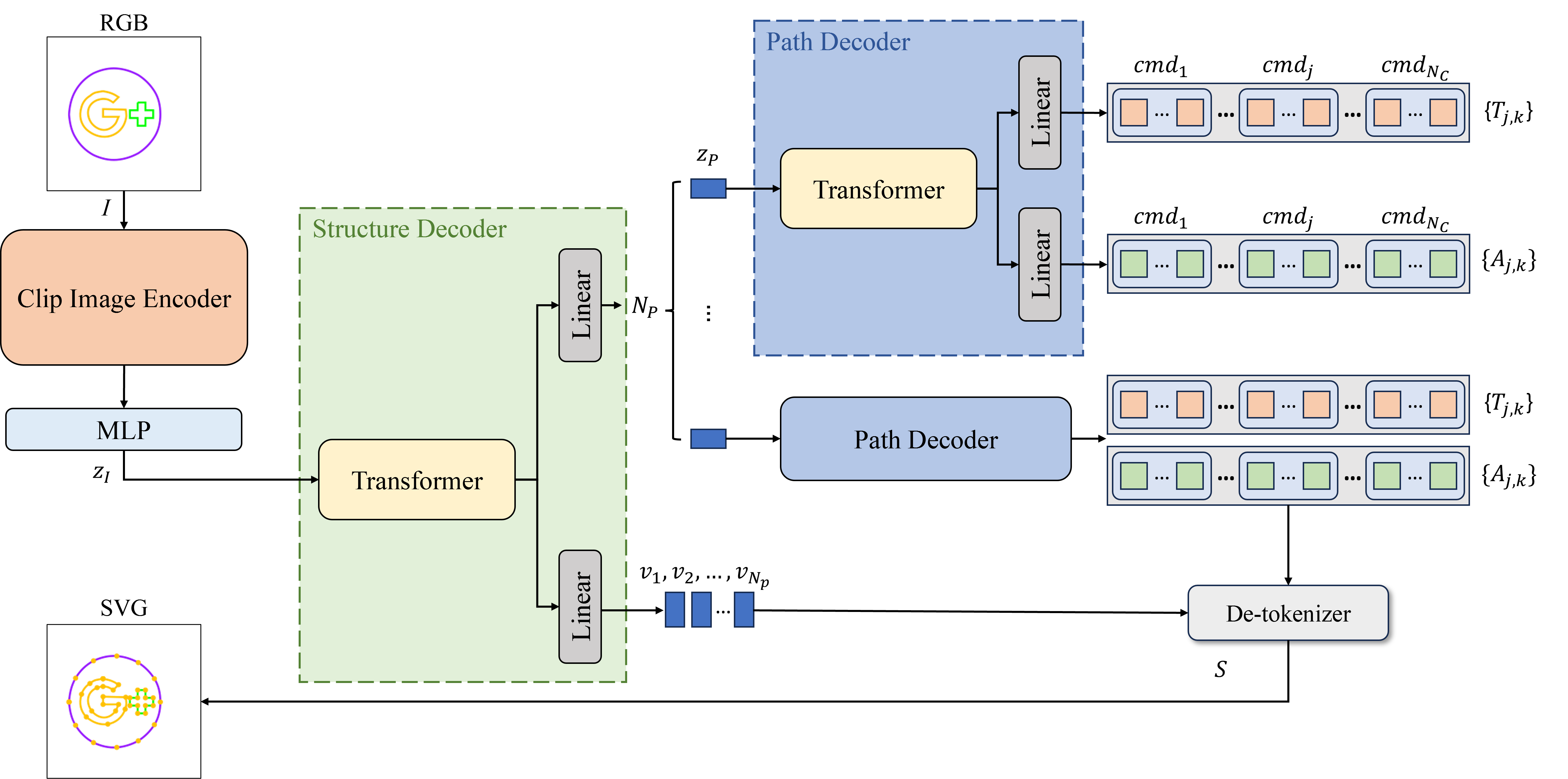}
  \caption{The overall architecture of DeepIcon. The input image is encoded with a CLIP Image Encoder~\cite{Clip} to generate the latent embedding $z_I$. Then, it will be fed into a Structure Decoder to infer a series of path embeddings $z_P$ and corresponding path visibility attributes $v_P$. For each path embedding, an individual path decoder outputs a pair of sequences $(\{T_{j,k}\}, \{A_{j,k}\})$ that defines the attributes and continuous arguments for each inferred SVG path. Here we use $T_{j,k}$ and $A_{j,k}$ to indicate the $k^{th}$ tokens from the $j^{th}$ command sequences within path $P_i$.
  }
  \label{fig:network}
\end{figure*}

\subsection{Model Architecture}
\label{sec:archi}
We develop a hierarchical deep architecture, as illustrated in Fig.~\ref{fig:network}, designed to efficiently generate high-quality SVG scripts from a single input image. The illustrated model can be generally segmented into three principal modules: an image encoding module (CLIP Image Encoder), a structure decoding module (Transformer Decoder), and a path decoding module (Transformer Decoder). The input image is initially processed through the CLIP Image Encoder, augmented by a trainable Multi-layer Perceptron (MLP), to produce a singular image embedding, denoted as $z_I$. 
We adopt CLIP as the encoder to better capture the semantic information in the input image. Since the IMG-to-SVG task requires the model to bridge the image context with SVG scripts, which are written in an XML-based language, CLIP is well-suited due to its strength in joint image-text embedding.
Drawing inspiration from DeepSVG~\cite{DeepSVG}, this image embedding is subsequently input into a hierarchical SVG decoder, which operates in two distinct stages: 1) The structure decoder outputs a latent representation $z_P$ and a corresponding visibility attribute $v$ for each path $P$). Following this, the path decoder translates each path embedding $z_P$ into a pair of tokens, as described in Section~\ref{sec:tokenization}.
\subsubsection{Structure Decoder}
The structure decoder within our architecture is comprised of a 4-layer Transformer decoder featuring a feed-forward dimension of $512$ and a model dimension of $256$. This transformer-based decoder is subsequently coupled with two trainable linear layers tasked with generating the path embeddings $z_P=(z_1, ..., z_{N_P})$ and the visibility attributes $v=(v_1, ..., v_{N_P})$ for each individual path $P_i$. Here $v_i\in \{0, 1\}$ serves as an indicator of path visibility. In this process, a predetermined number ($N_P=8$ in this paper) of path and visibility pairs, denoted as $(z_{P,i}, v_i)$, are generated by iteratively feeding the image embedding $z_I$ into the structure decoder. This innovative approach to predicting path visibility enables our network to generate SVG paths of variable lengths directly from the single image embedding, without the necessity for additional information. 
\subsubsection{Path Decoder}
For the task of path decoding, we employ a $12$-layer Transformer configured identically to the structure decoder. Each path decoder is also followed by $2$ trainable linear layers designed to transform the output of the Transformer into two distinct sequences. The first sequence, a type sequence denoted as $\{T_{j,k}\}$, has a dimension of $6$, corresponding to six different types of tokens. The second sequence, an argument sequence represented as $\{A_{j,k}\}$, contains continuous values. The continuous values $A_{j,k}$ in the argument sequence are only referenced when the predicted token $T_{j,k}$ in the type sequence is $\mathtt{arg}$.  

\subsection{Training}
\label{sec:training}
Since we treat the SVG scripts as a pair of sequences, the primary training objective of our model is to minimize the three distinct loss components: the cross-entropy loss for types $\{T_{j,k}\}$, the cross-entropy loss for the path visibility attributes $v_i$ and the mean squared error (MSE) for the valid arguments $\{A_{j,k}\}$. 
% For a pair of sequences $(\{T_{j,k}\}, \{A_{j,k}\})$, we can represent them as $\{T_{j,k}\}=[t_1, t_2, ..., t_{N}]$ and $\{A_{j,k}\}=[a_1, a_2, ..., a_{N}]$ for ease of explanation, where $N$ is the fixed length of padded sequences for each path. 
Therefore, for each path $P_i$, our objective is to optimize the model parameters $\Theta$ by minimizing the combined loss across these components:

\begin{align}
  \ell^{type}_{i} & = \frac{1}{N}\sum^N CE(\hat{T_{j,k}}, T_{j,k} | I; \Theta) {,} \\
  \ell^{vis}_{i} & = CE(\hat{v_i}, v_i | I; \Theta) {,} \\
  % \ell^{args}_{i} & = MSE(\hat{A_{j,k}}, A_{j,k} | I; \Theta) {,}
  \ell^{args}_{i} & = \frac{1}{N}\sum^N SE(\hat{A_{j,k}}, A_{j,k} | I; \Theta) {,}
  % \label{eq:losses_0}
\end{align}

\noindent where $N$ is the length of padded sequences for each path. $CE(\cdot, \cdot)$ denotes the cross-entropy function and $SE(\cdot, \cdot)$ refers to the squared error function, respectively. The symbol $I$ represents the input image to the image vectorization model. The triplets $(\hat{T_{j,k}}, \hat{A_{j,k}}, \hat{v_i})$ and $(T_{j,k}, A_{j,k}, v_i)$ correspond to the predicted and ground-truth attributes for each path $P_i$.
Consequently, the final loss function for our model is constructed as a weighted sum of these $3$ aforementioned losses, formalized as follows:
\begin{equation}
    \ell (\hat{P_i}, P_i) = \omega_{vis} \ell^{vis}_{i} + v_i \cdot (\omega_{type} \ell^{type}_{i} + \omega_{args}\textbf{1}_{T=\mathtt{arg}} \ell^{args}_{i}) {,}
    \label{eq:losses_1}
\end{equation}
\begin{equation}
    L(\hat{S}, S) = \sum_{i=1}^{N_P} \ell (\hat{P_i}, P_i) {,}
    \label{eq:losses_2}
\end{equation}
where $\hat{S}$ and $S$ denote the predicted SVG scripts and the target SVG scripts respectively. The weights for $\ell^{vis}$, $\ell^{type}$ and $\ell^{args}$ are denoted by $\omega_{vis}$, $\omega_{type}$, and $\omega_{args}$. In this paper, we have assigned the values $\ell^{vis}=1$, $\ell^{type}=1$ and $\ell^{args}=6\times 10^3$ for training, effectively calibrating the emphasis on argument accuracy due to its critical role in ensuring geometric precision in the generated SVG paths.
\subsubsection{Decoder Pretraining}
We first pretrain the SVG decoders (i.e., the structure decoder and the path decoder) through an SVG-to-SVG reconstruction task, utilizing the SVG encoder module as designed by DeepSVG. The SVG coordinates are quantized to 8 bits and are individually embedded for input into the SVG encoder. Additionally, the types of commands are converted into embeddings of the same dimension using a learned matrix. The final input is thus a composite of the type embedding, argument embedding, and an additional learned index embedding. At this stage, the training objective is to minimize a loss function similar to that described in  \eqref{eq:losses_2}, without considering the input $I$:
\begin{equation}
    L(\hat{S}, S | S, \Theta) = \sum_{i=1}^{N_P} \ell (\hat{P_i}, P_i) {.}
\end{equation}
% We utilize the AdamW optimizer~\cite{adamw} with an initial learning rate of $5\times 10^{-4}$ incorporating a linear warm-up over 500 steps. The pretraining phase lasts approximately 4 days, utilizing a batch size of 60 across 3 RTX 3090 GPUs.
\subsubsection{Joint Training}
Following the pretraining phase, we proceed to jointly finetune the CLIP image encoder and the pretrained SVG decoders for an IMG-to-SVG generation task. This is achieved by integrating a trainable 3-layer Multilayer Perceptron (MLP) head to bridge the two components. Here we utilize a pretrained CLIP model with a ViT-L/14 Transformer architecture from the Hugging Face community. The loss function employed at this stage is consistent with that described in \eqref{eq:losses_2}, thus can be written as:
\begin{equation}
    L(\hat{S}, S | I, \Theta) = \sum_{i=1}^{N_P} \ell (\hat{P_i}, P_i) {.}
\end{equation}
% For this phase, we adopted a reduced learning rate of $10^{-6}$ and the model was trained over a span of approximately 4 days, with a batch size of 20. 

\section{Experiments and Results}
\label{sec:experiments}
In this section, we begin with an introduction to data preparation, as detailed in Section~\ref{sec:data}. Subsequently, we evaluate the performance of our proposed approach by conducting comparisons with state-of-the-art (SOTA) methods, as outlined in (Section~\ref{sec:SOTA}), and by presenting ablation studies in Section~\ref{sec:ablation}.
\subsection{Data Preprocessing}
\label{sec:data}
We evaluated our method using the publicly available SVG-Icons8 dataset, which was processed as described by \cite{DeepSVG}. This dataset includes a diverse collection of 100,000 SVG icons, categorized across 56 distinct classes. To tailor the dataset to our network's requirements, we implemented a filtering process to exclude icons that were deemed unsuitable for our experimental setup. Specifically, icons comprising more than 8 paths or those containing paths with more than 32 commands were excluded from consideration. Following this criterion, the refined dataset consisted of 25,990 SVG icons. These were subsequently divided into two distinct subsets for the purposes of training and evaluating respectively: 18,193 icons constituted the training set, and 7,797 icons were designated for evaluation. This partitioning was executed randomly to ensure a representative distribution of icon categories across both sets.

\subsection{Implementation Details}
We utilize the AdamW optimizer~\cite{adamw} for both pretaining and joint training processes. For decoder pretraining, we use the optimizer with an initial learning rate of $5\times 10^{-4}$ incorporating a linear warm-up over 500 steps. The pretraining phase lasts approximately 4 days, utilizing a batch size of 60 across 3 RTX 3090 GPUs. For joint training, we adopted a reduced learning rate of $10^{-6}$ and the model was trained over a span of approximately 4 days, with a batch size of 20.

\subsection{Comparison with the State-of-the-art Methods}
\label{sec:SOTA}
\begin{table}[t]
\begin{center}
\caption{Inference comparison results on the evaluation set. }
\vspace*{1mm}
\label{tab:SOTA}
\begin{tabular}{lccccc}
  \hline
  \multirow{2}{*}{Model}  & \multicolumn{2}{c}{SVG-to-SVG} &  \multicolumn{2}{c}{IMG-to-SVG}   \\
\ & $IoU$ ($\uparrow$)& CD ($\downarrow$)& $IoU$ ($\uparrow$)& CD ($\downarrow$)&
  \\
  \hline
  LIVE~\cite{LIVE} & - & - & \textbf{0.2513} & -\\
  % \hline
  DeepSVG~\cite{DeepSVG} & 0.3543 & 0.0257  & 0.2049 & 0.0399\\
  DeepIcon (ours) & \textbf{0.3730} & \textbf{0.0238} & 0.2203 & \textbf{0.0377}\\
  \hline
\end{tabular}
\end{center}
\end{table}
We compared the performance of our proposed model with two other leading state-of-the-art (SOTA) image vectorization and SVG generation methods: LIVE and DeepSVG. Given that DeepSVG's primary application is the generation of SVG from the latent representations, we adapted it for direct comparison by finetuning the pretrained DeepSVG decoder in conjunction with the CLIP image encoder. The finetuning is employed in a similiar procedure as that used for DeepIcon. The quantitative outcomes of this comparative analysis are presented in Table~\ref{tab:SOTA}. For the purposes of this quantitative comparison, we employed two evaluation metrics: Intersection over Union ($IoU$) and Chamfer Distance (CD), for their relevance and reliability in assessing the accuracy and quality of vector graphics.
\subsubsection{Quantitative Analysis}
Table~\ref{tab:SOTA} presents the quantitative results comparing our proposed model, DeepIcon, with SOTA vectorization methods, specifically LIVE and DeepSVG, across two distinct tasks: SVG-to-SVG reconstruction and IMG-to-SVG vectorization. The results indicate that DeepIcon outperforms DeepSVG in both tasks, demonstrating superior SVG generation accuracy. It is important to note that LIVE does not include a SVG encoder, thus becoming not applicable for the SVG-to-SVG reconstruction task. Therefore, this comparison is exclusively between DeepSVG and DeepIcon. In this task, DeepIcon exhibited higher average $IoU$ and lower CD across the evaluation set, signaling more effective reconstruction performance with SVG data.

In the domain of IMG-to-SVG vectorization, DeepIcon maintains its superior performance over DeepSVG. When it comes to image vectorization, the SVG decoder of DeepIcon still performs better than DeepSVG. However, it is noteworthy that the $IoU$ for DeepIcon is slightly lower than that of LIVE. This discrepancy can be attributed to LIVE's approach of optimizing a stack of layered SVG paths through a differentiable rasterizer, which inherently may offer advantages in specific contexts. As discussed in Fig.~\ref{fig:example}, methods that prioritize image fitting, such as LIVE, tend to achieve better results on pixel-based metrics.
\begin{figure}[tb]
  \centering
  \begin{subfigure}{0.48\linewidth}
    % \fbox{\rule{0pt}{0.5in} \rule{.9\linewidth}{0pt}}
    \includegraphics[width=\linewidth]{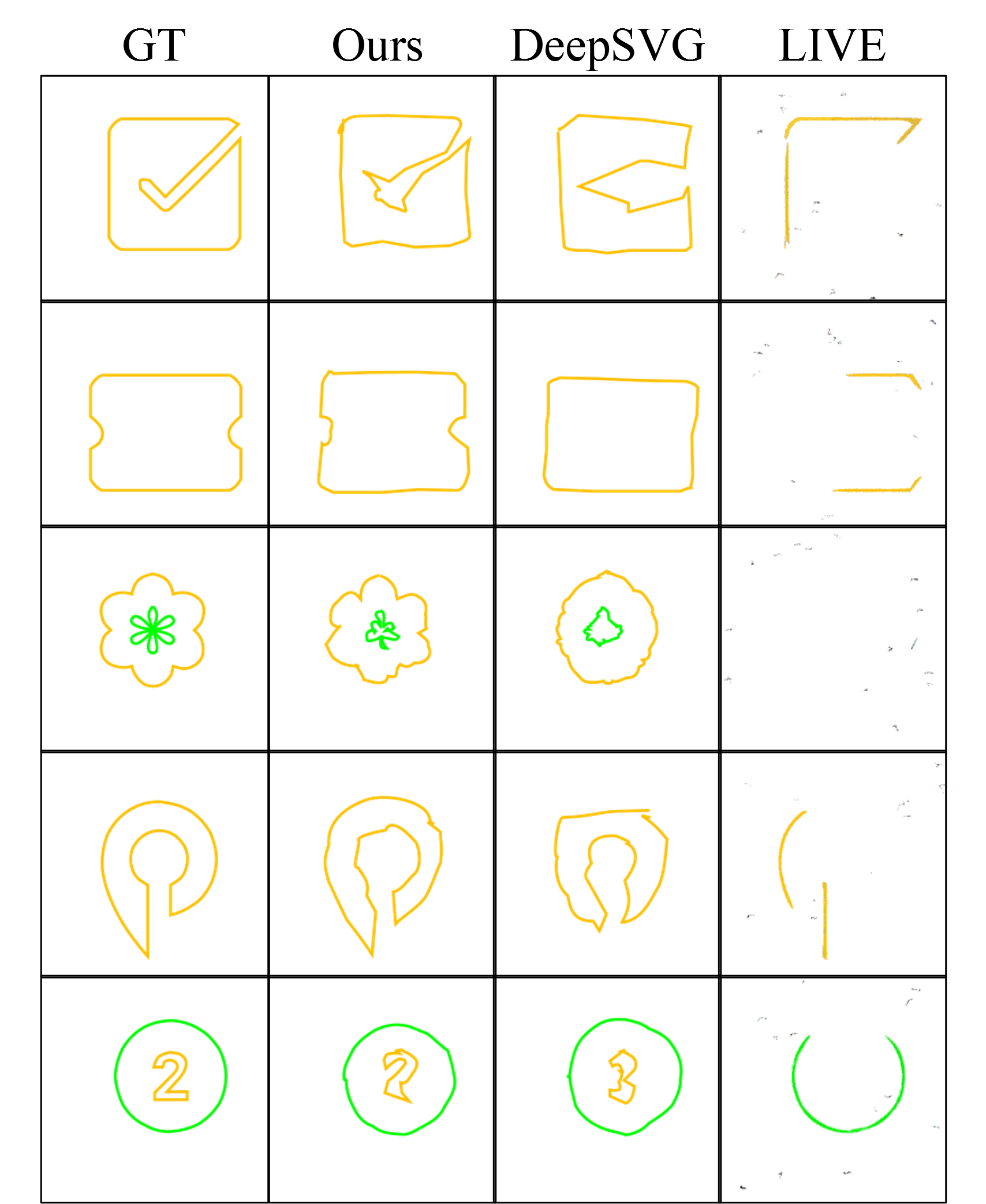}
    \caption{}
    \label{fig:sota-a}
  \end{subfigure}
  \hfill
  \begin{subfigure}{0.48\linewidth}
    \includegraphics[width=\linewidth]{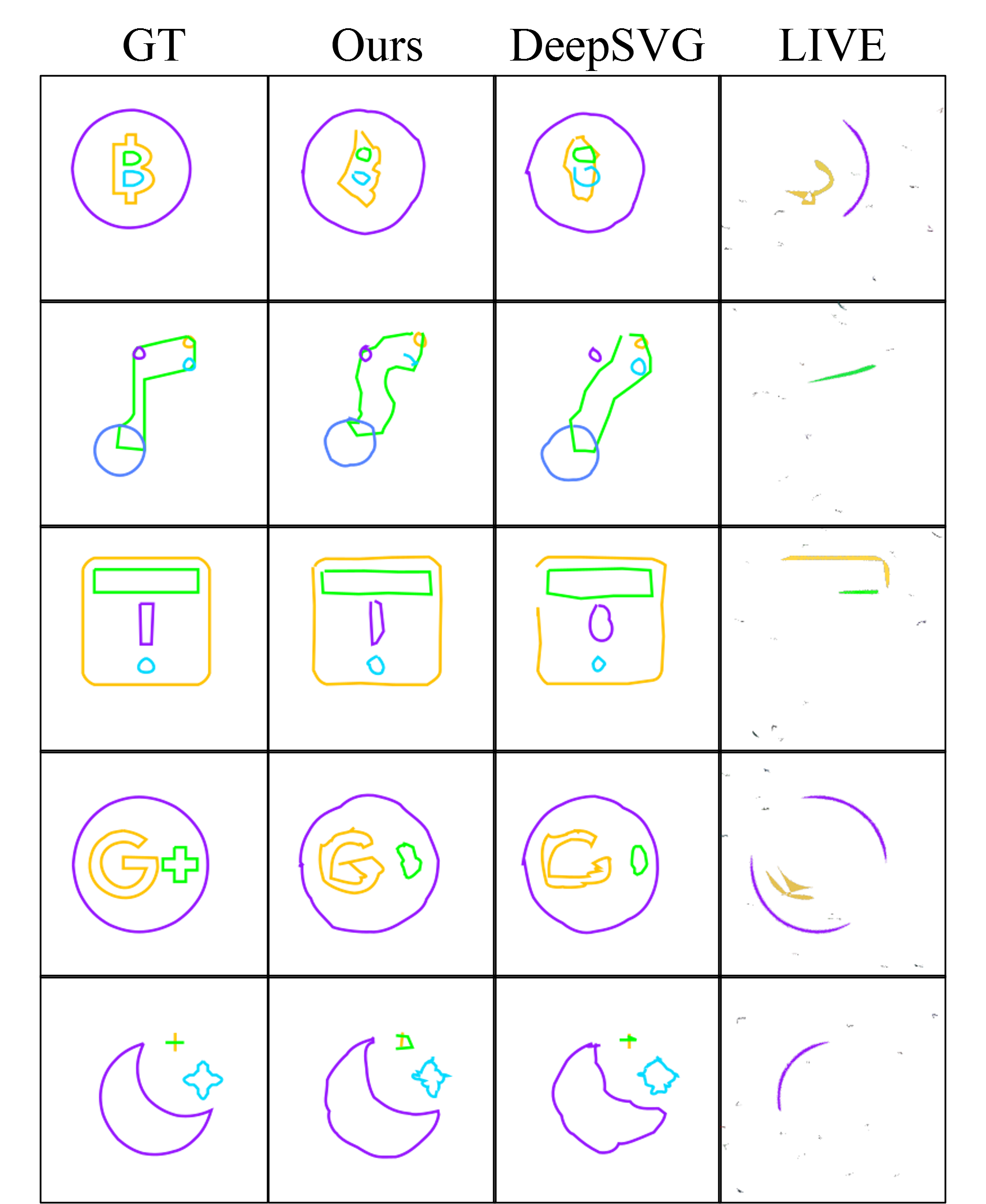}
    % \fbox{\rule{0pt}{0.5in} \rule{.9\linewidth}{0pt}}
    \caption{}
    \label{fig:sota-b}
  \end{subfigure}
  \caption{Our qualitative comparison with SOTA methods DeepSVG and LIVE.}
  \label{fig:SOTA}
\end{figure}
\subsubsection{Qualitative Analysis}
Fig.~\ref{fig:SOTA} presents a qualitative comparison among DeepIcon, DeepSVG, and LIVE for the IMG-to-SVG task. It is evident that DeepIcon surpasses the other methods in preserving the structure of complex shapes, as illustrated in Fig.~\ref{fig:sota-a}. Furthermore, DeepIcon demonstrates a superior understanding of the relationships between overlapping shapes, enabling it to generate SVG paths with greater accuracy in terms of preserving the content semantics and relations between shapes. Although LIVE achieves the highest $IoU$ as documented in Table~\ref{tab:SOTA}, it produces the least visually appealing results, as depicted in the figures. This outcome stems from the inherent limitations of parameter optimization methods, which, while adept at closely matching pixel values, often overlook higher-level information such as path relationships and the optimal number of paths. Consequently, LIVE tends to generate redundant but incomplete paths and is susceptible to falling into local optima.

\subsection{Ablations}
\label{sec:ablation}
\begin{table*}[t]
\begin{center}
\caption{Quantitative results of ablation study over the IMG-to-SVG generation task. The `Finetune' column denotes whether or not to finetune the pretrained CLIP image encoder. Additionally, we explore the implementation of a non-autoregressive decoder, which is based on the design principles outlined by \cite{DeepSVG}. Detailed explanations of these design choices and their implications are provided in Section~\ref{sec:ablation}.}
\vspace*{1mm}
\label{tab:ablat}
\begin{tabular*}{0.7\textwidth}{@{\extracolsep{\fill}} c|ccc|cc}
\hline
Model & Finetune  & Decoder            & Args       & IoU ($\uparrow$)             & CD ($\downarrow$)             \\ \hline
A     & \Checkmark                             & Non-autoregressive & Discrete   & 0.2049          & 0.0399          \\ \hline
B     & \XSolidBrush                           & Autoregressive     & Discrete   & 0.0669          & 0.1110          \\ \hline
C     & \Checkmark                           & Autoregressive     & Discrete   & 0.1868          & 0.0442          \\ \hline
D     & \XSolidBrush                         & Autoregressive     & Continuous & 0.0818          & 0.0951          \\ \hline
E     & \Checkmark                           & Autoregressive     & Continuous & \textbf{0.2203} & \textbf{0.0377} \\ \hline
\end{tabular*}
\end{center}
\end{table*}

\begin{figure}[tb]
  \centering
  \includegraphics[width=1\linewidth]{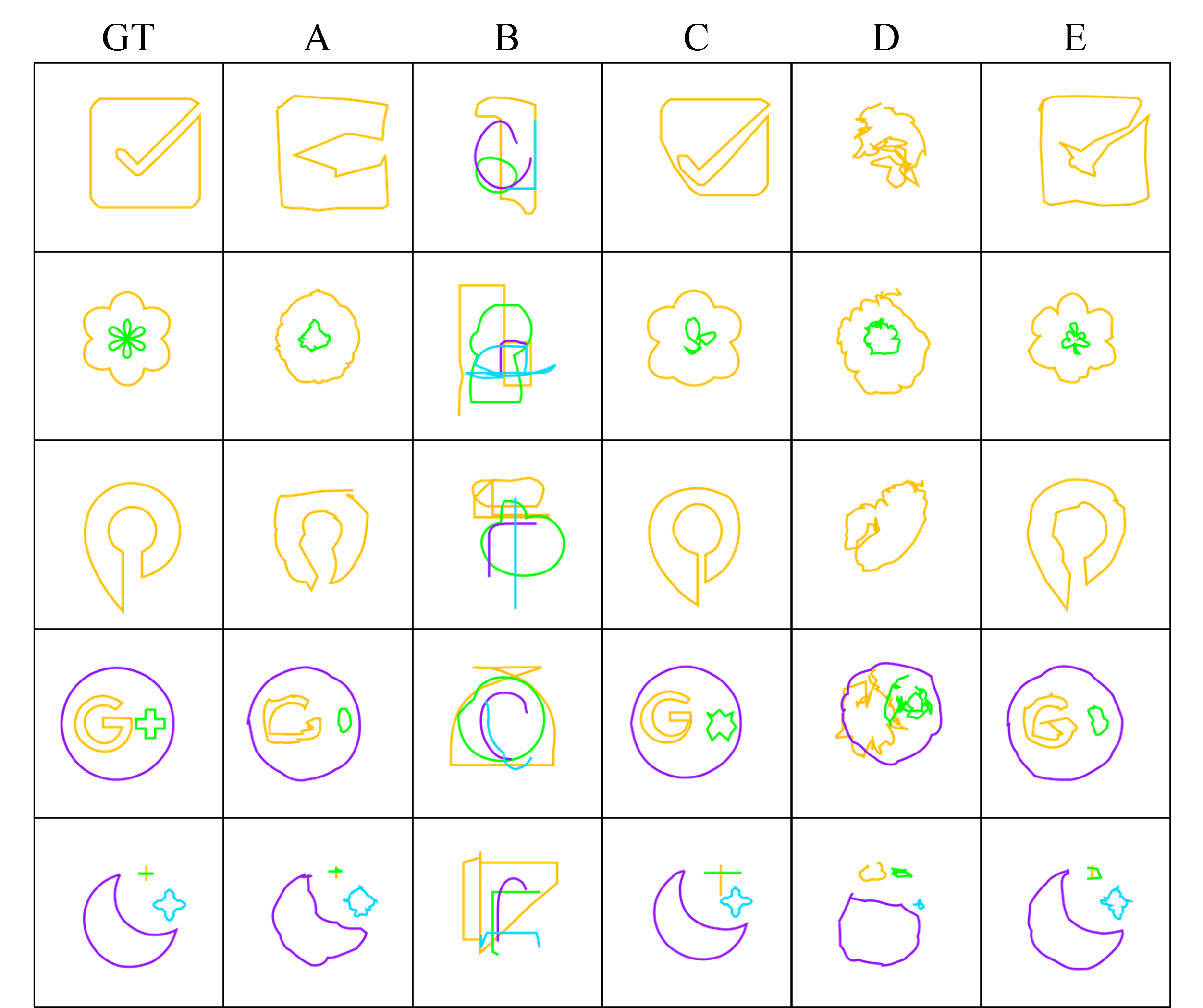}
  \caption{Ablation study on our proposed image vectorization network. The first column showcases the target image, which also serves as the input for our models. The specific configurations for models A through E are detailed in Table~\ref{tab:ablat}.
  }
  \label{fig:ablat}
\end{figure}

We conducted an ablation study on the evaluation set to validate the essential components of DeepIcon. By default, our configuration includes finetuning a pretrained decoder and a 3-layer MLP head, which is positioned after the CLIP image encoder to map the output embeddings to the input dimension required by the transformer. This study examines the impact of employing an autoregressive transformer, utilizing a continuous argument sequence, and applying finetuning techniques on the same dataset division. The outcomes of this investigation, including both quantitative and qualitative comparisons, are detailed in Table~\ref{tab:ablat} and Fig.~\ref{fig:ablat}.

\subsubsection{Effectiveness of Autoregressive Transformer}
In our evaluation, we first analyze the impact of employing an autoregressive transformer compared to a non-autoregressive approach in the path decoder's design. The non-autoregressive decoder projects the transformer's output dimension directly to the product of the number of arguments in a command and the argument embedding dimension, denoted as $n_{args}\times d_{args}$, where $n_{args}=11$.  Unused arguments within the command sequence are padded with $-1$. Consequently, the non-autoregressive decoder processes path embeddings as inputs and predicts tokens by the unit of commands. As demonstrated in Model A and C of Table~\ref{tab:ablat} and Fig.~\ref{fig:ablat}, employing a non-autoregressive transformer decoder leads to a noticeable decline in accuracy, as evidenced by the lower $IoU$ and higher CD. However, autoregressive decoder provides apparently better visual performance, yielding much clearer shapes delineations and superior semantics consistency. This suggests that while autoregressive decoders are adept at capturing and reproducing the semantic essence of shapes, they struggle with precise geometric positioning. 

\subsubsection{Effectiveness of Continuous Arguments}
Then, we examine the impact of transitioning arguments into a continuous space for prediction. The details of the implementation are introduced in Section~\ref{sec:tokenization}. Given the observed trade-offs between autoregressive and non-autoregressive decoding approaches, our goal is to preserve the semantic understanding benefits offered by the autoregressive transformer while addressing its limitations in terms of coordinate inference accuracy. By shifting to a continuous space, we observe significant improvements, both quantitatively and qualitatively. Predicting arguments in the continuous space not only enhances the accuracy of coordinate predictions but also maintains the autoregressive transformer's advantage in capturing the semantics of shapes.

\subsubsection{Effectiveness of Finetuning}
While most downstream applications employ the CLIP image encoder as a zero-shot encoder with an additional MLP (Multilayer Perceptron) head for training, we compare the outcomes with and without finetuning the CLIP image encoder, as depicted in Table~\ref{tab:ablat} and Fig.~\ref{fig:ablat}. Without finetuning, the discrete space SVG (Scalable Vector Graphics) decoder struggles to learn effectively from the examples, resulting in relatively random SVG outputs, as illustrated in the column B of Fig.~\ref{fig:ablat}. This issue is partially addressed by adopting continuous arguments, as demonstrated in the column D of Fig.~\ref{fig:ablat}. However, the results still fall short of those achieved with a finetuned CLIP encoder.

\section{Conclusion}

This paper introduces a novel image vectorization network designed to generate variable-length SVG scripts from a single image input, reconstructing its structure with an emphasis on semantic consistency. DeepIcon reconceptualizes SVG icon representation as a sequence of paths, wherein each path consists of a variable number of commands delineating parametric lines or curves, thereby prioritizing the completion of topological structures. The model is trained and evaluated on the SVG-Icons8 dataset. Utilizing the CLIP image encoder and the hierarchical transformer-based SVG decoder, our network effectively bridges the gap between images and SVG scripts. The introduction of a unique tokenization rule and the alignment of continuous arguments further enhance the model's performance. The quantitative and qualitative results from our experiments demonstrate that DeepIcon surpasses state-of-the-art approaches, achieving high-quality icon vectorization results.

\textbf{Limitation.} While the DeepIcon model demonstrates achievements in the domain of icon vectorization, it remains improvable. While the reconstruction accuracy of the icon vectorization process is not yet perfect, it suggests an opportunity for the development of a more sophisticated SVG decoder, aimed at significantly enhance SVG decoding performance. Also, DeepIcon is limited to predicting relatively simple SVG shapes, with a maximum of 8 paths and no more than 32 commands per path.

\textbf{Future Works.} Enhancing the capabilities of the SVG decoder could not only improve the model's accuracy but also reduce its reliance on the finetuning of the CLIP image encoder. Such advancements could pave the way for the model's extension into new applications, including text-to-SVG generation. By potentially substituting the image encoder with a pretrained CLIP text model, DeepIcon could be adapted to interpret and translate textual descriptions into SVG scripts, broadening its applicability and utility in the field of graphic representation.

\bibliographystyle{IEEEtran}
\bibliography{egbib}
\end{document}